\documentclass[conference, a4paper]{IEEEtran}
\IEEEoverridecommandlockouts
\usepackage{cite}
\usepackage{amsmath,amssymb,amsfonts}
\usepackage{algorithmic}
\usepackage{graphicx}
\usepackage{textcomp}
\usepackage{xcolor}
\usepackage{subcaption}
\usepackage{multirow}
\usepackage{booktabs}
\usepackage{hyperref}
\def\BibTeX{{\rm B\kern-.05em{\sc i\kern-.025em b}\kern-.08em
    T\kern-.1667em\lower.7ex\hbox{E}\kern-.125emX}}
\begin{document}

\title{Multivariate Time Series Anomaly Detection with Few Positive Samples\\
}

\author{
	\IEEEauthorblockN{Feng Xue}
\IEEEauthorblockA{\textit{GE Research} \\
	Niskayuna, NY, USA \\
	xue@ge.com}
\and
	\IEEEauthorblockN{Weizhong Yan}
\IEEEauthorblockA{\textit{GE Research} \\
	Niskayuna, NY, USA \\
	yan@ge.com}
}

\maketitle

\begin{abstract}
Given the scarcity of anomalies in real-world applications, the majority of literature has been focusing on modeling 
normality. The learned representations enable anomaly detection as the normality model is trained to capture
certain key underlying data regularities under normal circumstances. In practical settings, particularly industrial 
time series 
anomaly detection, we often encounter situations where  a large amount of normal operation data is available along 
with a small number of anomaly events collected 
over time. This practical situation calls for methodologies to leverage these small number of anomaly events to create 
a better anomaly detector. In this paper, we introduce two methodologies to address the needs of this practical 
situation and compared 
them with recently developed state of the art techniques. Our proposed methods anchor on representative learning of 
normal operation with autoregressive (AR) model along with loss components to encourage representations that 
separate normal versus few positive examples. 
We applied the proposed methods to two industrial anomaly detection datasets and demonstrated effective 
performance in comparison with approaches from literature. Our study also points out additional challenges with 
adopting such methods in practical applications. 

\end{abstract}


\section{Introduction}
Anomaly detection has been a widely researched topic in machine learning and is of significant importance in
many areas such as fraud detection, cyber security, and complex system health monitoring
\cite{chandola_anomaly_2009}. It
still remains as an active and challenging research area. In recent years, deep learning has been widely used for
anomaly detection
\cite{chalapathy_deep_2019}. Anomaly detection has been applied to many different applications, 
this paper mainly focuses on  its application to industrial multivariate time series data. With the increasing number of 
sensors as well as 
cost-effective data transmission and storage solutions, industrial systems, such as power plant, wind turbines, 
engines etc., produce large amounts of time series data during their regular operations. It is important to monitor 
these 
systems to spot 
abnormal behaviors, which, if not detected earlier, could have significant reliability consequences.

Anomalies, also referred to as outliers, are defined as observations which deviate so much from the majority of all the
observations. In the context of industrial time series data, the systems are usually operated based on what they are
designed for. Under normal operating conditions (NOCs), the system measurements are a partial capture of the 
dynamic state governed by operation profiles, first principles, and the underlying control logic. Hence, the contextual 
information is an important factor when developing anomaly detection techniques. These anomalies are  often 
referred as  contextual anomalies, one of  anomaly types as 
categorized in \cite{chandola2009anomaly}. 

For a typical anomaly detection application, there is usually a lack of abnormal data. Most of the data collected for anomaly
detection model development is under NOCs. The core idea of developing anomaly detection is
to learn the spatial (cross multiple system measurement and commands) and temporal relationships under normal
operations. In an abnormal situation, such relationships will not adhere to the learned representation, resulting in
deviations  from the normal operating patterns. The higher a deviation is, the more likely it is an anomaly. This type of 
anomaly detection methods, i.e., to learn the system normal behabior using data under NOCs only, is often referred as  
semi-supervised anomaly detection (SSAD) in the literature. 

In a real-world application setting, some labeled anomalies (faults or events) are sometimes available (albeit the 
number of such 
anomalies is usually very small), in addition to the abundant availability of normal data. Leveraging such 
limited anomaly data in the process of building anomaly detection algorithms to improve their detection 
performance, mainly 
reducing false positives, has became an interesting research question. In recent years, a few research efforts have 
been 
made towards answering this very research question.  For example, Pang et al. \cite{PangWeaklysupervisedAD} 
proposed the Pairwise Relation 
prediction-based ordinal regression Network, which simultaneously learns pairwise relations and anomaly 
scores by training an end-to-end ordinal regression neural network.

In this paper, we explore the use of loss functions based on few anomalous samples to regulate learning normality 
representation, hence increase the model's effectiveness of detecting anomalies. The contributions of this paper are as 
follows:

\begin{itemize}
	\item We described a jointly-learning approach that incorporates both normality representation learning and 
	regularization from few anomalous samples.
	\item We demonstrated advantages of such learning strategy over state of the art methodology on two public 
	datasets. 
	\item We further examined that the behavior of applying such learning approach to situations where available 
	anomalous samples are not representative of future anomalies, i.e. a domain shift, and revealed the limitation of the 
	current approaches towards real-world applications. 
\end{itemize}

\section{Related Work}
\subsection{Normal Time Series Data Modeling}
In the process control and model-based fault detection community, a number of data driven approaches have been
used for anomaly detection of industrial time series data, for example, Principal Component Analysis (PCA),
Dynamic PCA
\cite{ku_disturbance_1995}, subspace aided approach \cite{ding_subspace_2009}. These traditional approaches 
take into consideration of multivariate linear relationship, and to some extent of temporal dependence in the case of 
Dynamic PCA or subspace
aided approach. Another often used approach is one-class SVM such as in~\cite{ma_time-series_2003}. In recent 
years, deep
learning has become an active research area for multivariate anomaly detection \cite{malhotra_lstm-based_2016,
	hundman_detecting_2018, malhotra_lstm-based_2016, guo_multidimensional_2018, zong_deep_2018, 
	su_robust_2019,
	xue_deep_2020}.

A number of approaches have been proposed  in the literature for modeling normal time series data. One class is to 
learn an autoregressive (AR) model, in which the past observations are used to predict the future. A Recurrent 
Neural Network (RNN) such as
a LSTM~\cite{hochreiter_long_1997} is usually
used for such tasks, although recent research work \cite{oord_wavenet_2016, bai_empirical_2018} demonstrated 
that a causal convolutional
network might be a better alternative in term of effectiveness and training efficiency. Another
class is the encoder-decoder based approach. For example, \cite{malhotra_lstm-based_2016} is a direct application 
of sequence-to-sequence (seq2seq)
modeling as in \cite{cho_learning_2014} with reconstruction errors as the loss function for time series
representation learning. Alternatively in \cite{park_multimodal_2018}, a variational autoencoder (VAE) has been
employed along with an LSTM to
reconstruct each input at each time step of the series. \cite{guo_multidimensional_2018} proposed GGM-VAE, a
Gaussian Mixture model that is used
to represent the latent space with GRU as an encoder. GGM-VAE aims to better deal with multimodal sensory data, in
contrast to a typical single Gaussian latent space representation. In our problem setting, we have both normal data 
and few faulty samples. The goal is to find a better normality representation by leveraging both for enhancing anomaly 
detection effectiveness. 

\subsection{Semi-Supervised Anomaly Detection}
As discussed in \cite{villa2021semi}, traditional semi-supervised anomaly detection (SSAD) methods involve 
training  models using labeled normal samples only and ignoring limited labeled anomaly samples available, 
which tends to have a high false positive rate. Recently developed SSAD methods focus on improving anomaly 
detection performance, i.e., reducing false positives, by leveraging the labeled anomaly data (often very small). These 
new SSAD methods differ primarily in the strategies used for leveraging the labeled anomaly samples, depending on 
data  availability scenarios and the problem settings.

Assuming both labeled and unlabeled samples are available, Ruff, et al. \cite{ruff2019deep} proposed DeepSAD, an 
end-to-end methodology for general SSAD, which is an extension of DeepSVDD \cite{ruff2018deep}. DeepSVDD is a 
neural network version of support vector data description (SVDD) with a specially defined objective function such 
that it can learn feature representation and the smallest hypersphere together. While DeepSVDD works on unlabeled 
data and assumes most of the unlabeled data are normal, DeepSAD takes advantage of labeled samples in addition to 
unlabeled data. Essentially it includes an additional term in the DeepSVDD’s cost function to force the network to map 
normal samples closer to the hypersphere center and the known anomalies further from it. 

Considering an application scenario where only a small labeled anomaly data and a large number of unlabeled data are 
available, Pang et al. \cite{PangWeaklysupervisedAD} proposed the Pairwise Relation prediction-based ordinal 
regression Network, which simultaneously learns pairwise relations and anomaly scores by training an 
end-to-end ordinal regression neural network. They termed their anomaly detection problem as “weakly-supervised 
anomaly detection”. By addressing the same application scenarios, \cite{ADOA_Zhang} proposed an anomaly 
detection with partially observed anomalies. More recently, this type of anomaly detection problem was 
tackled in \cite{gao2021connet} where a SSAD method called ConNet was proposed. It utilized a few labeled 
anomalies as a prior knowledge and trained an anomaly scoring network with the concentration loss, an improved 
version of the contrastive loss. 

Our work in this paper is also related to leveraging few labeled anomaly samples for improving anomaly detection 
performance. However, our work differs significantly from the aforementioned studies in that we are addressing 
anomaly detection problem in an industrial setting where the data is primarily time-series sensor measurements. 
Similarly to DeepSAD, our proposed method uses additional loss term derived from labeled anomalies in comparison 
to 
normal data, but our normal representative learning leverage time series data temporal property instead of a mere 
projection to a hypersphere as in DeepSAD. In our paper, we compared our approach with DeepSAD  on two 
benchmark datasets.

\begin{figure*}[ht]
	\centering
	\begin{subfigure}[t]{0.33\linewidth}
		\centering
		\includegraphics[width=0.9\linewidth]{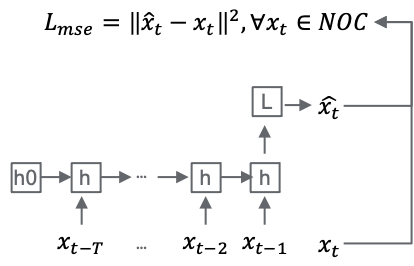}
		\caption{AR model}
		\label{fig:model_ar}
	\end{subfigure}%
	\begin{subfigure}[t]{0.33\linewidth}
		\centering
		\includegraphics[width=0.9\linewidth]{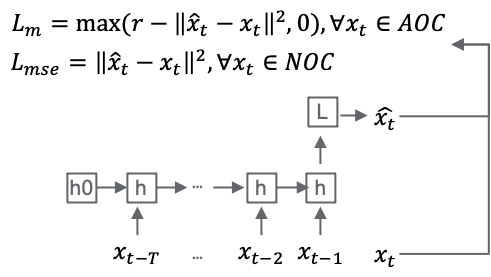}
		\caption{AR model with margin loss}
		\label{fig:model_margin}
	\end{subfigure}%
	\begin{subfigure}[t]{0.33\linewidth}
		\centering
		\includegraphics[width=0.9\linewidth]{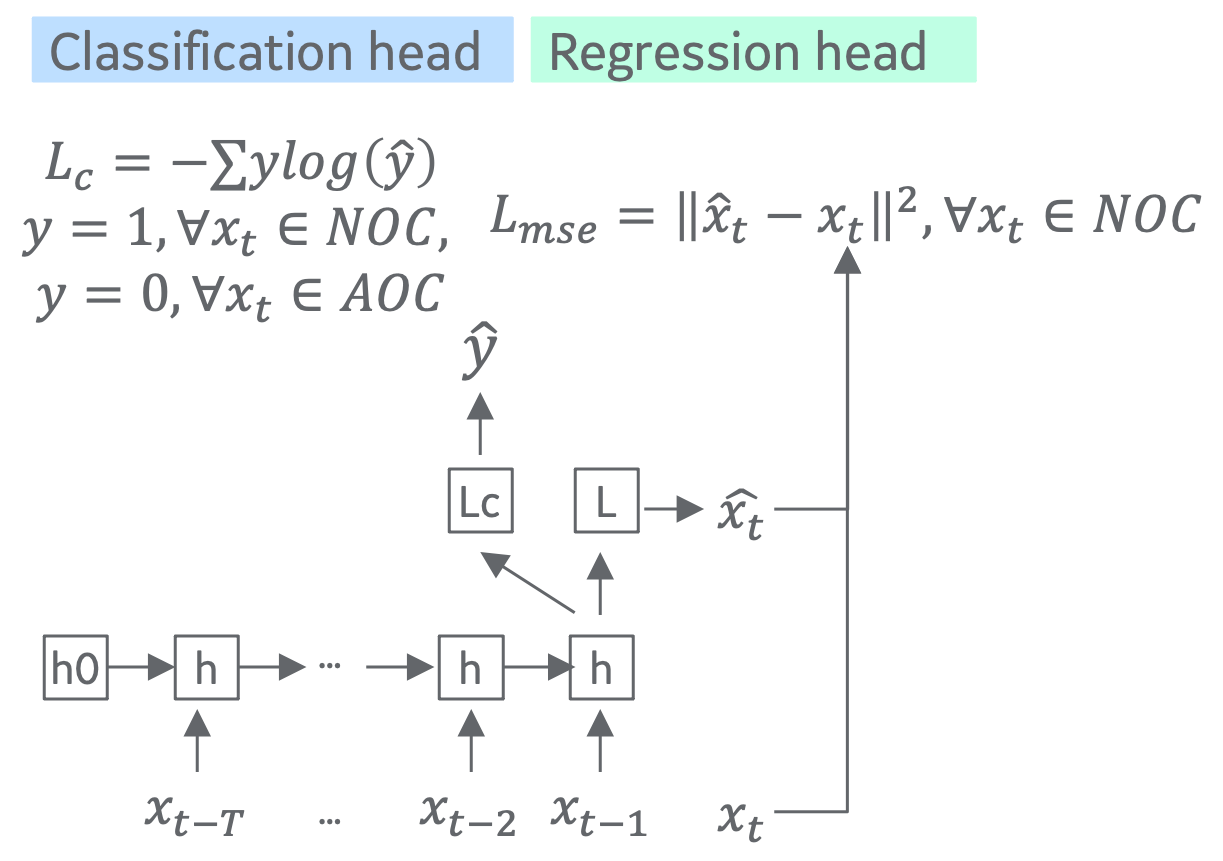}
		\caption{AR model with auxiliary classification loss}
		\label{fig:model_auxiliary}
	\end{subfigure}
	\caption{Comparing traditional AR model with the proposed approach for handling few fault samples. 
		\ref{fig:model_ar} 
		illustrates a traditional AR model with a recurrent network (such as LSTM) as backbone; \ref{fig:model_margin} 
		illustrates the margin loss approach with the same backbone; while \ref{fig:model_auxiliary} illustrated the 
		auxiliary classification task approach with the same backbone}
	\label{fig:model_formulation}
\end{figure*}

\section{Problem Formulation and Approach}~\label{sec:problem_formulation}
In general, anomaly detection with neural networks can be categorized into three paradigms: deep learning for
feature extraction, learning feature representation of normality, and end-to-end anomaly score
learning~\cite{pang2020deep}. In a typical industrial setting, normal operation data are usually abundant while the
number of faulty cases is often very small if there are any. Therefore, the second category, which models the 
normality, is often a preferable approach. This paradigm learns a representation of data by using an objective function 
that is not directly aligned to an anomaly score. 

When it comes to modeling industrial time series data, an autoregressive (AR) approach is often used for modeling 
such time 
series given its connection to dynamic systems. Let 
$\mathbf{x_t}\in \mathbb{R}^n$ be the multivariate sample of 
dimension $n$ at time $t$, and denote the $j$-th dimension at time $t$ as $x^j_t$ (i.e., $\mathbf{x_t}=[x_t^1, 
x_t^2,...,x_t^n]$). The AR approach is trying to estimate $\mathbf{x_t}$ from all observations up to time $t-1$, 
noted as $\mathbf{x_{t-}}$, which indicates all time series data before $t$. 
That is to say, the model is trying to learn the relationship $\mathbf{\hat{x}_t}=f(\mathbf{x_{t-}})$, such that the 
prediction $\mathbf{\hat{x}_t}$ is as close to the real observation 
$\mathbf{x_t}$ as possible measured by some distance metric $d(\mathbf{\hat{x}_t, x_t})$. 
In practice, a window of length $T$ is often used as the inputs to the model instead of all samples prior to time $t$. 
This window length can be adjusted to different applications and datasets. The distance metric $d$ can be chosen as 
the Euclidean distance, corresponding to a mean squared error (MSE) loss during training as in 
Equation~\ref{eq:mse_ar}: 
\begin{equation}\label{eq:mse_ar}
	L_{mse} = \|\mathbf{\hat{x}_t - x_t} \|^2
\end{equation}
This distance metric measures the deviation of a sample from what it should have been under normal operating 
conditions. Therefore, a sample whose deviation is above a defined threshold can be regarded as anomalous.
Under the assumption of that majority of the time series data gathered in industrial settings is under NOC, this 
training mechanism essentially tries to learn the dynamic representation of the underlying system. In the case that 
there are a number of known anomalies, we have majority time series $\mathbf{x_t} \in \mathcal{N}$ from 
NOC, while minority time series $\mathbf{x_t} \in \mathcal{A}$ from Anomalous Operation Conditions (AOC). 
In order to leverage these samples from AOC, we formulated two approaches: 1) MSE margin loss; 2) auxiliary 
classification task. The proposed approaches are illustrated in Figure \ref{fig:model_formulation}.

\subsection{MSE Margin Loss}
In this formulation, we want to encourage network to not only reduce the MSE loss $L_{mse}$ in Equation 
\ref{eq:mse_ar} for samples from NOC, but also produce a higher MSE for samples from AOC. To that end, we  
introduce a margin loss as following: 
\begin{equation} \label{eq:loss_m}
	L_{m} = \max(r - \|\mathbf{\hat{x}_t - x_t} \|^2, 0) , \forall x_t \in \mathcal{A} 
\end{equation}
where $r$ is $p^{th}$ percentile of MSEs from the NOC samples. In practice, $r$ is the exponential moving average 
of $p^{th}$ percentile of MSEs from NOC samples in each batch during training. Here the loss term 
penalizes the AOC samples with MSE less than the margin $r$. The overall loss is: 
\begin{equation} \label{eq:loss_m_all}
L = L_{mse} + \alpha L_{m} 
\end{equation}
where $\alpha$ is a hyper parameter. 
\subsection{Auxiliary Classification Task }
In this formulation, we have an auxiliary classification task in addition to the main AR task. In this case, the input 
$\mathbf{x_{t-}}$ is transformed in to hidden state via $\mathbf{z}=h(\mathbf{x_{t-}, W_h})$ (the last hidden 
state as illustrated in \ref{fig:model_auxiliary}). From here, one 
branch of the network aims to produce AR output, i.e. $\mathbf{\hat{x}_t}=g(\mathbf{z, W_g})$; while the other 
branch 
of the network aims to map the hidden state to a binary class output  $\mathbf{\hat{y}}=o(\mathbf{z, W_o})$. 
Hence, a 
cross-entropy loss can be used to encourage the network to distinguish NOC samples versus AOC samples. 
\begin{equation} \label{eq:loss_c}
L_{c} = \sum_{i=0}^{1}(y_i log(\hat{y}_i))
\end{equation}
where $y_i$ is the $i$-th component of  the one-hot  vector $\mathbf{y}$ of the true class label $(0, 1)$ 
representing normal or anomalous samples, while $\mathbf{\hat{y}}$ is the estimated vector. 
Similarly, the overall loss is: 
\begin{equation} \label{eq:loss_c_all}
L = L_{mse} + \alpha L_{c} 
\end{equation}
where $\alpha$ is a hyper parameter. 

\section{Experimental Results}
\subsection{Datasets}
\subsubsection{TEP Dataset}\footnote{TEP dataset can be downloaded at 
\url{https://dataverse.harvard.edu/dataset.xhtml?persistentId=doi:10.7910/DVN/6C3JR1}.}
The Tennessee Eastman process (TEP) is an industrial benchmark by the Eastman Chemical Company for process 
monitoring and control studies~\cite{downs1993plant}. It models a real industrial process computationally and is 
widely studied for anomaly detection algorithms~\cite{yin2012comparison, sun2020fault}.

The TEP is comprised of $4$ reactants, $2$ products, $1$ by-product and $1$ inert components denoted as A-H. 
These components undergo a chemical process enabled by 5 major units: a reactor where the reaction happens for the 
gas feed components (A, C, D and E) into liquid products (G and H), a condenser that  cools down the gas stream 
coming out of the 
reactor, a separator that separates gas and liquid components from the cooled product stream, a compressor that 
feeds the gas stream back into the reactor and a stripper that strips the two products from any unreacted feed 
components.

The TEP dataset~\cite{DVN/6C3JR1_2017} contains $52$ variables in total, $41$ of which are sensor 
measurements (XMEAS(1) - XMEAS(41)) and $11$ are manipulated variables ((XMV(1) - (XMV11)). Therefore, 
the multivariate samples 
$\mathbf{x_t}$ have a dimension of $52$, or $n=52$. The dataset has separate training and testing files, both of 
which contain a set of  ``fault-free'' and ``faulty'' files. Each file contains single simulation run of the chemical process.  
The 
``fault-free'' runs correspond to the processes under normal operating conditions (NOC) while the  ``faulty'' files 
contains 20 different simulated process faults. There are 500 simulation runs for both normal and each faulty 
operations. Each  test data run has a  length of  $960$ (representing 48 hours of 
operation 
sampled at 1/3min). For each faulty run in the test dataset, the first $160$ samples are under normal operation, with 
the remaining  $800$ samples under certain faulty condition. For fault detection rate (FDR) calculations below, we 
only considered the length $800$ faulty region. 
\begin{equation*}
	\mathrm{FDR=\frac{number \: of \: alarms \: in \: faulty \: region}{total  \:  samples \: in  
 \: faulty \: region}}
\end{equation*}
And the false alarm rate (FAR) is calculated as:
\begin{equation*}
	\mathrm{FAR=\frac{number \: of \: alarms \: in \: NOC}{total  \: samples \: in \: NOC}} 
\end{equation*}

It should be noted that in this dataset, controllable faults (Fault $3,9,15$) have disturbances that can be dealt with by
the control system, and therefore they return to normal regions. In these circumstance, the FDR is not expected to be
significantly different from the FAR~\cite{sun2020fault}.

In our experiment setup, we take a small number of runs from each fault type in the faulty training dataset along with 
all the $500$ runs under NOC. These faulty 
data are batched along with the fault free data in the training process . 

\subsubsection{HAI Dataset} \footnote{HAI dataset can be downloaded at 
\url{https://github.com/icsdataset/hai}.} 
The HIL-based Augmented ICS (HAI) Security Dataset was collected from a realistic industrial control system (ICS) 
testbed augmented with a Hardware-In-the-Loop (HIL) simulator that emulates steam-turbine power generation and 
pumped-storage hydro power generation~\cite{shin2020hai}. Both normal and abnormal behaviors for ICS anomaly 
detection are included in the dataset, with the abnormal one collected based on various attack scenarios with the six 
control loops in three different types of industrial devices.

The HAI testbed consists of four processes: Boiler Process (P1), Turbine Process (P2), Water-treatment Process (P3)
and HIL Simulation(P4). During normal operation, it is assumed that the operator operates the facility in a
routine manner, while abnormal behaviors occur when some of the parameters are outside the  normal range or are in
unexpected states due to attacks, malfunctions, and failures. The experiment in this paper  is conducted based on the 
$20.07$ version of
HAI dataset, which has a training and testing  dataset with $n=59$ process measurements to model. The data also
contains label information about whether there is an attack and where in the three processes.
There are a total of 177 hours of data in the training set and 123 hours of data in the test set.

In order to mimic the situation of having a few known anomaly cases during model training, we picked 3 primitive 
attack scenarios from each
processes as training data. These anomaly cases are not included in the subsequent testing stage for 
performance reporting purpose. In Table \ref{table:hai_cases} we listed the specific cases used as training data in our 
experiment  so 
readers can repeat the same setting for future studies. These cases are a subset of the 38 attacks in the $20.07$ 
version of HAI dataset \footnote{HAI data attack details: 
\url{https://github.com/icsdataset/hai/blob/master/hai_dataset_technical_details_v2.0.pdf}} 

\begin{table}[htb]
	\centering
	\caption{Few positive samples included in HAI training data}
	\begin{tabular}{c|c|c|c}
		\toprule
		ID & Process & Start Time & Duration(sec) \\
		\midrule
		A101	& P1	& 10/29/19 13:40	& 370
 \\
		A102	& P1	& 10/29/19 14:35	& 312
\\
		A103	& P1	& 10/29/19 15:45	& 868
\\
		A110	& P2	& 10/30/19 14:30	& 370
\\
		A113	& P2	& 10/31/19 8:42	& 348
\\
		A116	& P2	& 10/31/19 13:25	& 368 \\
		A112	& P3	& 10/30/19 16:33	& 154
\\
		A111	& P3	& 10/30/19 15:35	& 180 \\
		A203	& P3	& 11/1/19 11:23	& 180 \\
		\bottomrule
	\end{tabular}
	\label{table:hai_cases}
\end{table}

\subsection{Model Setups}
In all experiments, we train the models with Adam optimizer~\cite{kingma_adam_2015} with $\beta_1=0.9$, 
$\beta_2=0.999$, $\epsilon=10^{-8}$ and a learning rate of $0.001$. Model batch size is set to be $1000$ and 
number of  epochs is $100$. The metrics for calculating loss during training and deviations during anomaly 
detection is MSE. For 
both training sets, 80\% of the data was used for training and 20\% for validation.

Our base model is a $2$ layer LSTM  with hidden state dimension of $50$. This is followed by a linear layer that maps 
the LSTM output to the final output, which  has a dimension of $n=52$ 
for TEP and $n=59$ for HAI respectively. For DeepSAD, we use exactly the same model setup  to represent final 
output encoding for a fair comparison. 

For both TEP and HAI data, we take a window of length $20$ as the inputs to the AR models and length $1$ as the 
output.  That is to say, we use $[\mathbf{x_{t-20}}, \mathbf{x_{t-19}},...,\mathbf{x_{t-1}}]$ to estimate 
$\mathbf{x_{t}}$. 

For the proposed auxiliary classification approach (Auxiliary-LSTM in short), the auxiliary branch takes the LSTM 
output 
and performs a linear mapping to a dimension of $2$ for a binary classification setup. 

For the proposed MSE margin loss approach (Margin-LSTM in short), $p^{th}\mathrm{percentile} of 95$ is used for radius 
$r$ calculation from NOC samples in each 
batch, a momentum of $0.9$ is used for exponential moving average over batches to update $r$, which is initialized 
at $0$.

\begin{table*}[htb]
	\centering
	\caption{TEP dataset fault detection results. The FAR for the NOC is set to be 5\%, and the average FDR 
	(with standard deviation) over 10 random trials is
	reported for each of the 20 fault types. We compare the results obtained from our proposed approach (with 3 faulty 
	runs),  the normal data only LSTM model, and DeepSAD (with the same 3 faulty 
	runs). Highest FDR is  presented in bold.}
	\begin{tabular}{c|c|c|c|c}
		\toprule
		& Normal AR Model  &  \multicolumn{3}{c}{Normal data + Few Faults}   \\
		Fault  & LSTM & Auxiliary-LSTM & Margin-LSTM & DeepSAD \\
		\hline
		1 &	0.9972 (0.0001)	& 0.9969 (0.0001)	& \textbf{0.9973} (0.0003)	&	0.9927 (0.0008)
\\
		2	&	0.9816 (0.0004)	& \textbf{0.9884} (0.0004)	&	0.9831 (0.0008)	&	0.9856 (0.0010)
 \\
		3	&	0.0511 (0.0003)	&	0.0822 (0.0251)		& 0.0514 (0.0012)	&	\textbf{0.1091} (0.0291)
\\
		4	&	0.9993 (0.0003)	&	\textbf{0.9999} (0.0003)	&	\textbf{0.9999} (0.0001) 	&	0.9092 (0.2700)
\\
		5	&	0.2305 (0.0066)	&	0.9356 (0.1200)	&	\textbf{0.9991} (0.001)	&	0.9236 (0.2181)
\\
		6	&	\textbf{1.0000} (0.0000)&\textbf{1.0000} (0.0000) &	\textbf{1.0000} (0.0000) &0.9973 (0.0007)	 
		\\
		7	&	\textbf{1.0000} (0.0000)&\textbf{1.0000} (0.0000)&	\textbf{1.0000} (0.0000) &0.9861 (0.0325)
 
		\\
		8	&	0.9479 (0.0019)	&	\textbf{0.9765} (0.0007)	&	0.9633 (0.0028)	&	0.9643 (0.0118)
\\
		9	&	0.0517 (0.0001)	&	\textbf{0.0717} (0.0126)	&	0.0529 (0.0011)	&	0.0710 (0.0088)
 \\
		10	&	0.1364 (0.0129)	&	0.8418 (0.0572)	&	\textbf{0.8861} (0.0244)	&	0.7378 (0.1973)
\\
		11	&	0.7982 (0.0036)	&	\textbf{0.9818} (0.0141)	&	0.9091 (0.0198)	&	0.9081 (0.2254)
\\
		12	&	0.9671 (0.0016)	&	\textbf{0.9914} (0.0002)	&	0.9858 (0.0018)	&	0.9808 (0.0160)
 \\
		13	&	0.9333 (0.0012)	&	\textbf{0.9513} (0.0012)	&	0.942 (0.001)	&	0.9407 (0.0066)
\\
		14	&	0.9996 (0.0001)	&	0.9995 (0.0000)	&	\textbf{0.9997} (0.0000)	&	0.9797 (0.0510)
 \\
		15	&	0.0533 (0.0001)	&	0.0584 (0.0008)	&	0.0535 (0.0002)	&	\textbf{0.0623} (0.0038) \\
		16	&	0.1282 (0.0072)	&	0.8974 (0.0766)	&	\textbf{0.9341} (0.0198)	&	0.7220 (0.2385)
\\
		17	&	0.9397 (0.0113)	&	\textbf{0.9623} (0.0008)	&	0.9621 (0.0004)	&	0.9179 (0.1243)
\\
		18	&	0.937 (0.0002)	&	\textbf{0.9398} (0.0007)	&	0.9382 (0.0005)	&	0.9338 (0.0077)
\\
		19	&	0.2335 (0.0019)	&	\textbf{0.7275} (0.1635)	&	0.4533 (0.0601)	&	0.3040 (0.1255)
\\
		20	&	0.6105 (0.0265)	&	0.9007 (0.0420)	&	\textbf{0.9096} (0.0163)	&	0.8303 (0.2636)\\
		
		Average &  0.6498 & \textbf{0.8152} & 0.8110 &0. 6625 \\
		\bottomrule
	\end{tabular}
	\label{table:tep}
\end{table*}

For experimental results presented in this section, a simple grid search was performed to get the best parameter 
setting. For TEP, margin loss weight $\alpha = 0.5$, and auxiliary classification loss  weight $\alpha = 0.5$. A 
search for DeepSAD yield loss weight $\eta = 0.01$ (Equation 7 in \cite{ruff2019deep}, which has the same meaning 
as $\alpha$ here). For HAI, margin loss  weight 
$\alpha = 1.0$,  auxiliary classification loss weight $\alpha = 0.01$,  and DeepSAD loss weight $\eta = 0.001$.

\subsection{Results and Discussions}\label{sec:exp:results}
\subsubsection{TEP Dataset}
We report the FDR for all $20$ fault types at the FAR of $5\%$ and the overall performance for the TEP data. For 
HAI data, we report the overall performance based on both ROC
(Receiver Operating Characteristic) and PR (Precision-Recall) curves. Both AUC (Area Under
Curve) for the ROC curve and Average Precision (AP) are calculated as the overall performance measure. 

Results for TEP dataset are summarized in Table~\ref{table:tep}. We compare the fault detection results using our 
proposed approach with baseline normal data only model, and DeepSAD. The FAR
for the normal case is set to be 5\%, and the FDR are presented in this table. The FDR results in the table are the 
average (with standard deviation) 
over 10 random trial for each model. It can be seen that using as little as 3 faulty 
runs from each fault type in training can improve the original LSTM model significantly. Both Auxiliary-LSTM  and 
Margin-LSTM show comparable performance, although Auxiliary-LSTM shows a slight edge in this experimental 
result. Although DeepSAD also improved the anomaly detection performance compared with normal data only model, 
the improvement is  marginal. 

We also report the overall performance (ROC-AUC and AP) regardless fault types in Table \ref{table:tep_overall}. 
Overall, 
Auxiliary-LSTM and Margin-LSTM perform better than normal model only. It should be noted that normal model only has 
the lowest standard deviation, while Margin-LSTM is only slightly higher. On the other hand, DeepSAD has a much large 
variation among independent random trials. 

\begin{table}[htb]
	\centering
	\caption{TEP dataset anomaly detection results. We report average ROC AUC and AP (with standard deviation) for 
		each model. Higher values are in bold.}
	\begin{tabular}{c|c|c}
		\toprule
		Method & ROC AUC & AP \\
		\midrule
		LSTM & 0.8427 (0.0010)  &  \textbf{0.9898} (0.0001)\\
		Auxiliary-LSTM & \textbf{0.9133} (0.0073) &  0.9948 (0.0005)\\
		Margin-LSTM  & 0.9071 (0.0022) &  0.9945 (0.0001)\\
		DeepSAD & 0.8829 (0.0485) & 0.9926 (0.0036)\\
		\bottomrule
	\end{tabular}
	\vspace{-1mm}
	\label{table:tep_overall}
\end{table}

\subsubsection{HAI Dataset}
Results for HAI dataset are reported in Table~\ref{table:hai}. In a similar way, we compare results from   
normal data only model as baseline with Auxiliary-LSTM, Margin-LSTM, and DeepSAD. 
Similarly to the TEP study, we report both the ROC AUC and AP.

\begin{table}[htb]
	\centering
	\caption{HAI dataset anomaly detection results. We report average ROC AUC and AP (with standard deviation) for 
		each model. Higher values are in bold.}
	\begin{tabular}{c|c|c}
		\toprule
		Method & ROC AUC & AP \\
		\midrule
		LSTM & \textbf{0.7947} (0.0023)  &  0.4992 (0.0078)\\
		Auxiliary-LSTM & 0.7801(0.0234) &  0.4422 (0.0251)\\
		Margin-LSTM  & 0.7941 (0.0133) &  \textbf{0.5226} (0.0231)\\
		DeepSAD & 0.7020 (0.0356) & 0.3344 (0.1285)\\
		\bottomrule
	\end{tabular}
	\vspace{-1mm}
	\label{table:hai}
\end{table}

In this case, we did not observe a consistent advantage from the proposed few faults model or DeepSAD. Indeed, 
DeepSAD actually performed worse than normal data only model. Both Auxiliary-LSTM  and 
Margin-LSTM  performs very  
close to normal data only model, although Margin-LSTM has a better AP.

\begin{figure*}[ht]
	\centering
	\begin{subfigure}[t]{0.33\textwidth}
		\centering
		\includegraphics[width=0.9\linewidth]{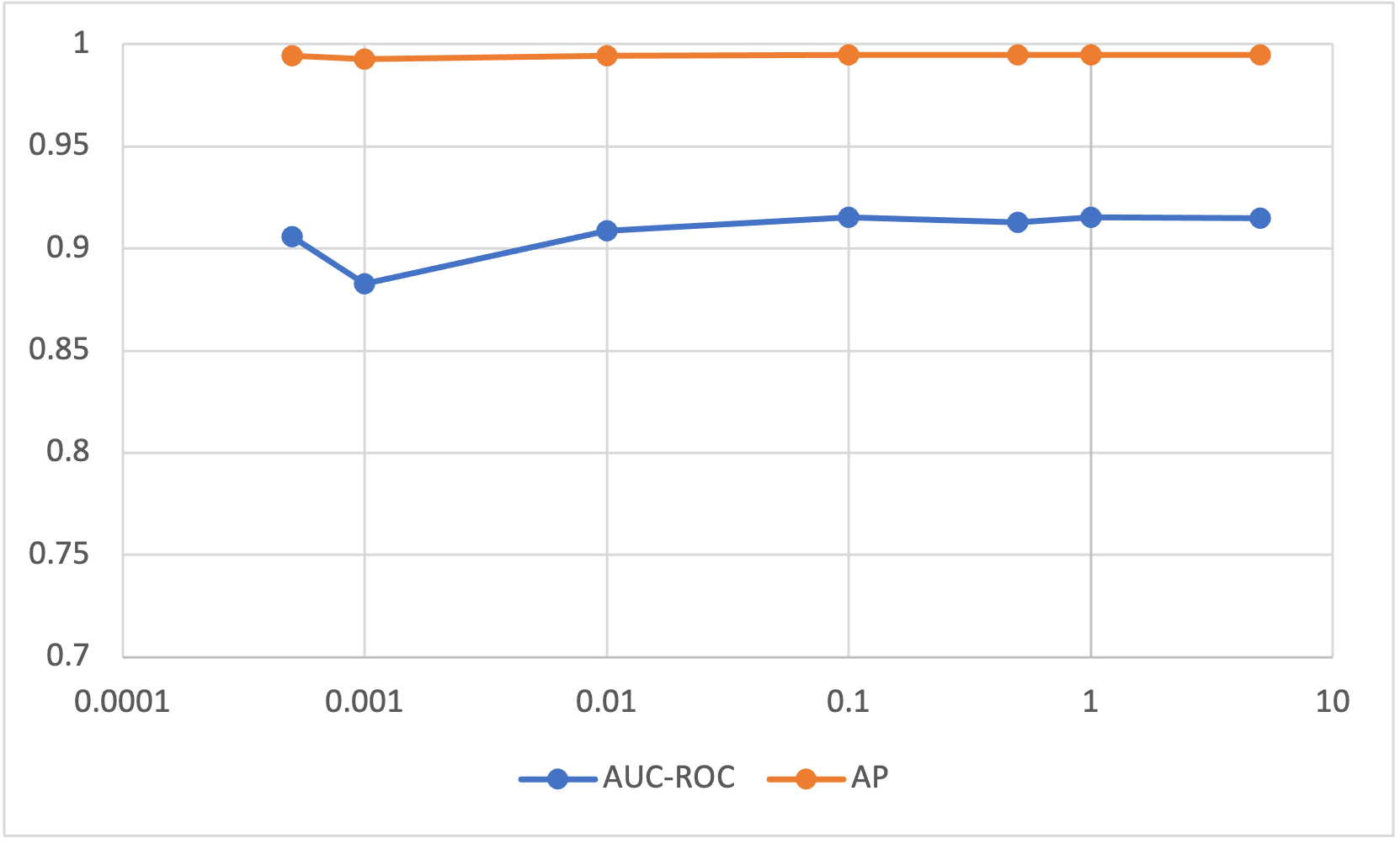}
		\caption{Auxiliary-LSTM}
		\label{fig:tep_auxiliary}
	\end{subfigure}%
	\begin{subfigure}[t]{0.33\textwidth}
		\centering
		\includegraphics[width=0.9\linewidth]{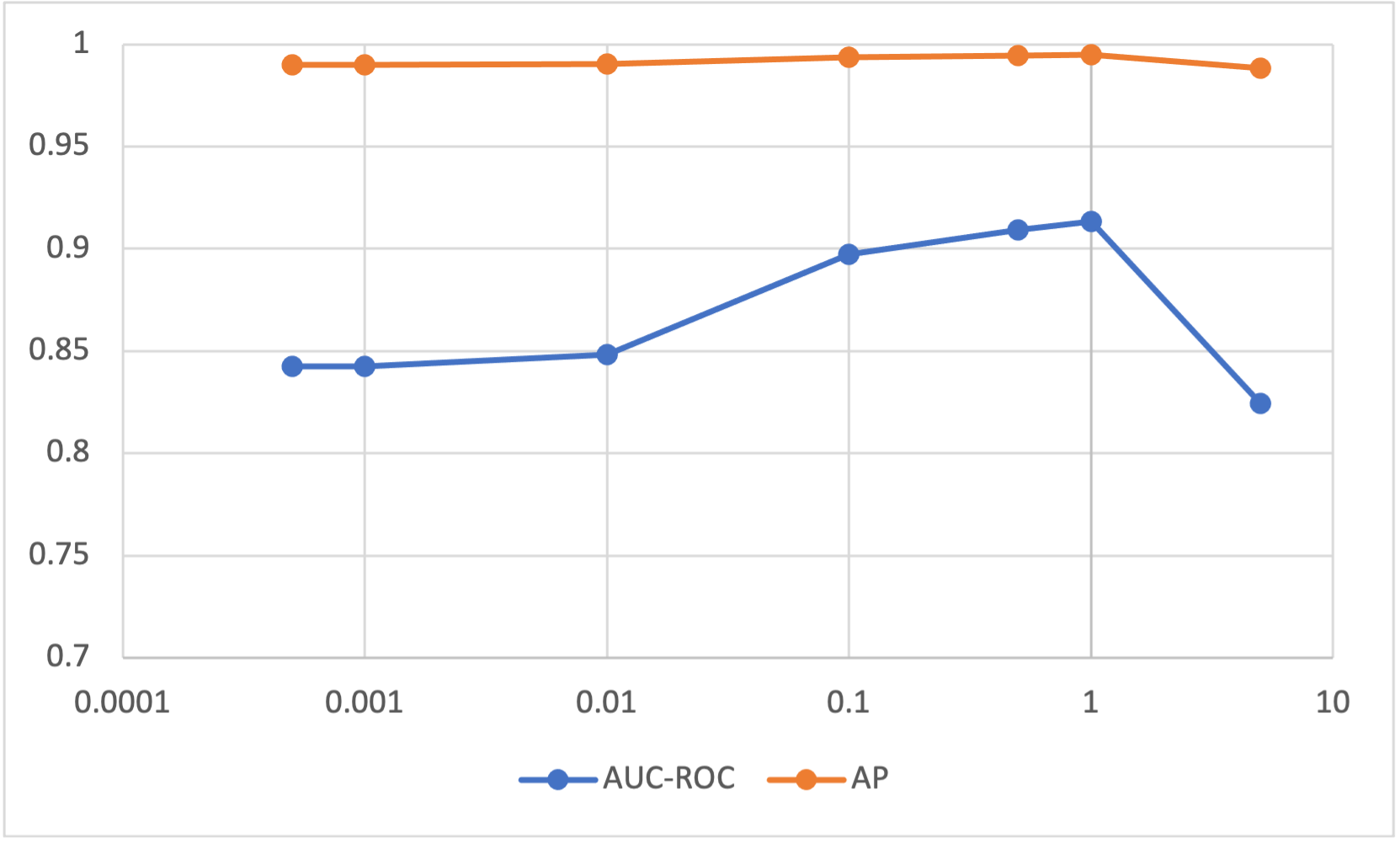}
		\caption{Margin-LSTM}
		\label{fig:tep_margin}
	\end{subfigure}%
	\begin{subfigure}[t]{0.33\textwidth}
		\centering
		\includegraphics[width=0.9\linewidth]{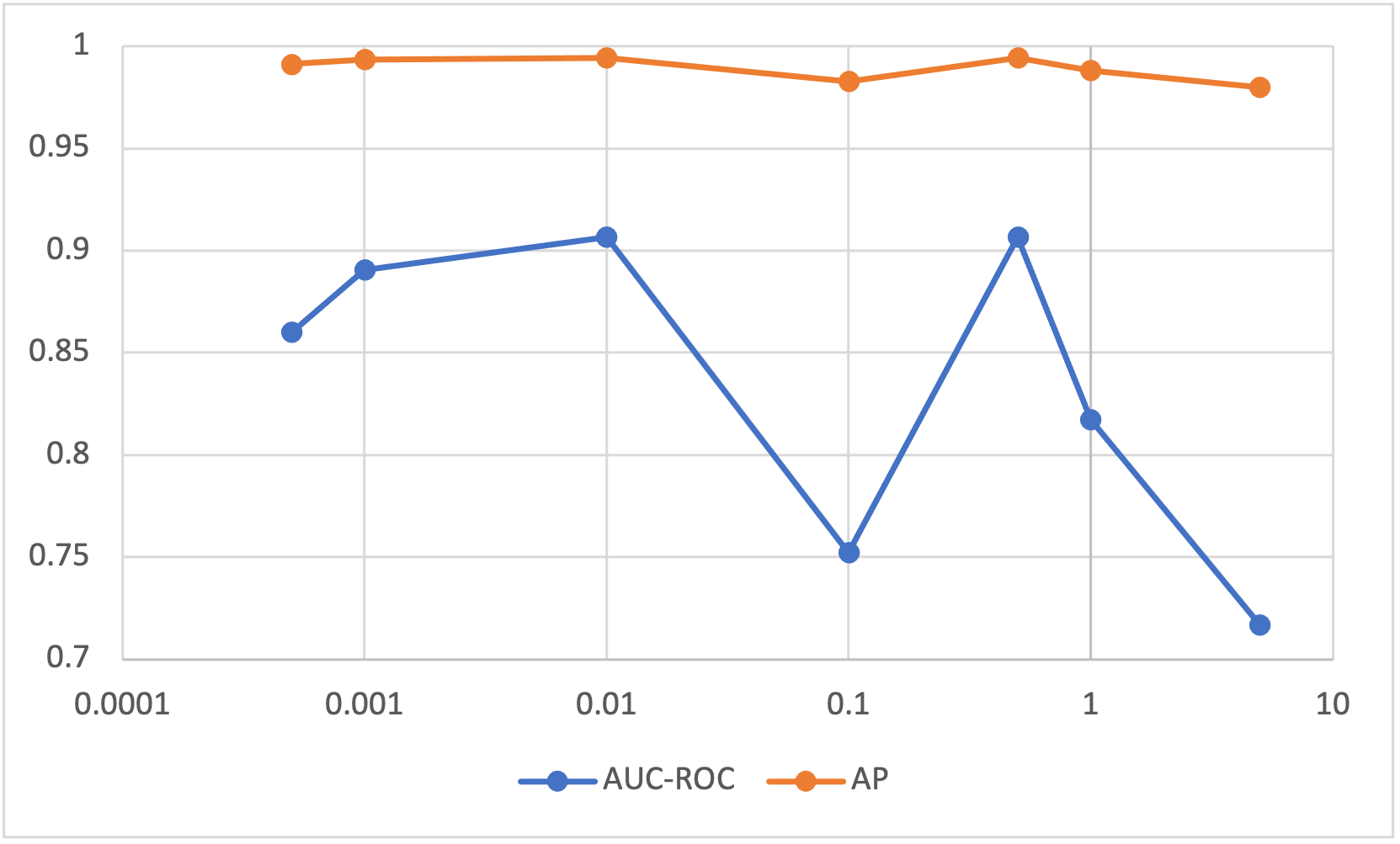}
		\caption{DeepSAD}
		\label{fig:tep_deepsad}
	\end{subfigure}
	\caption{Sensitivity of additional loss term weight  $\alpha$ on model performance with TEP data}
	\label{fig:sensitivity_tep}
\end{figure*}

\begin{figure*}[ht]
	\centering
	\begin{subfigure}[t]{0.33\textwidth}
		\centering
		\includegraphics[width=0.9\linewidth]{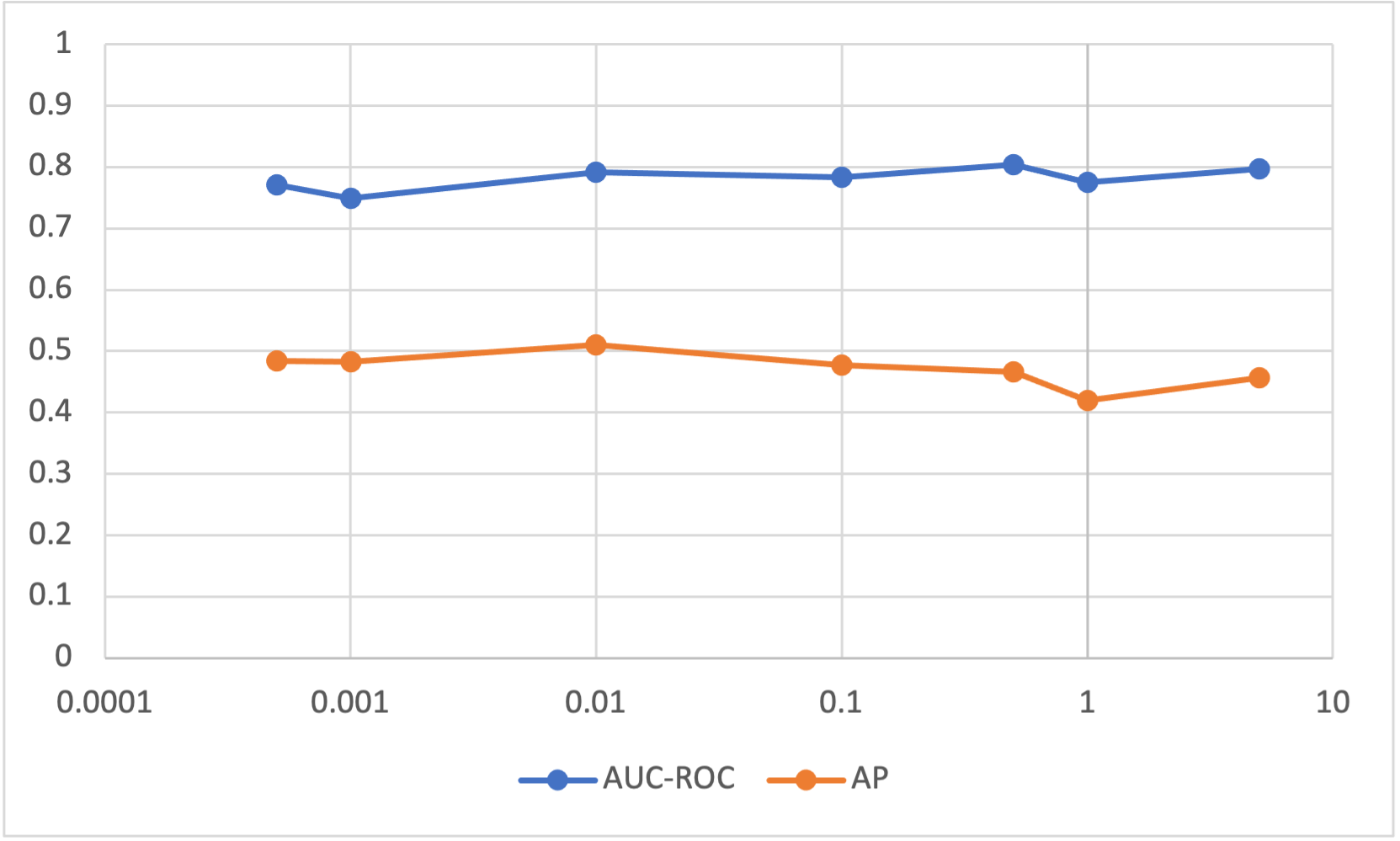}
		\caption{Auxiliary-LSTM}
		\label{fig:hai_auxiliary}
	\end{subfigure}
	\begin{subfigure}[t]{0.33\textwidth}
		\centering
		\includegraphics[width=0.9\linewidth]{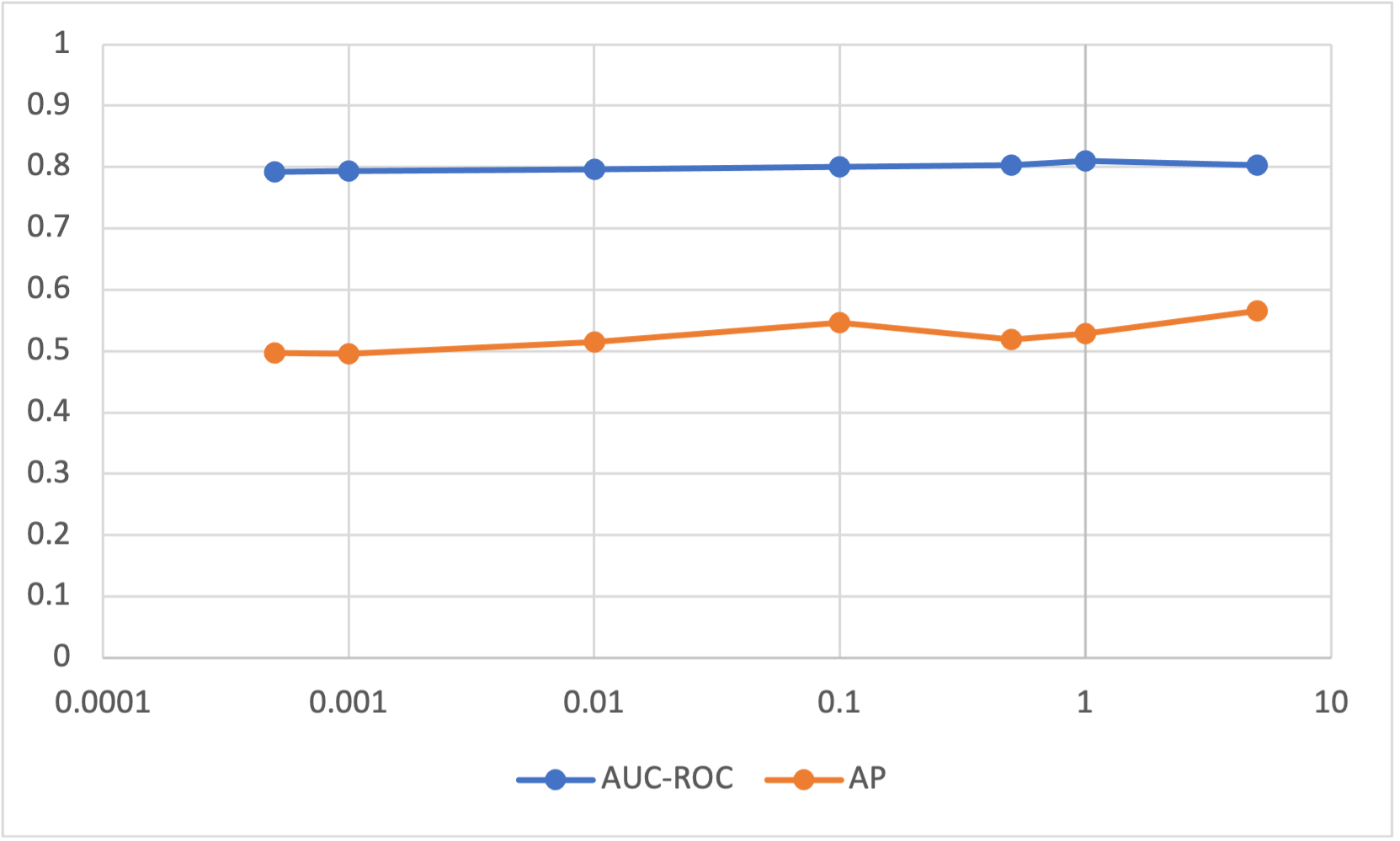}
		\caption{Margin-LSTM}
		\label{fig:hai_margin}
	\end{subfigure}
	\begin{subfigure}[t]{0.33\textwidth}
		\centering
		\includegraphics[width=0.9\linewidth]{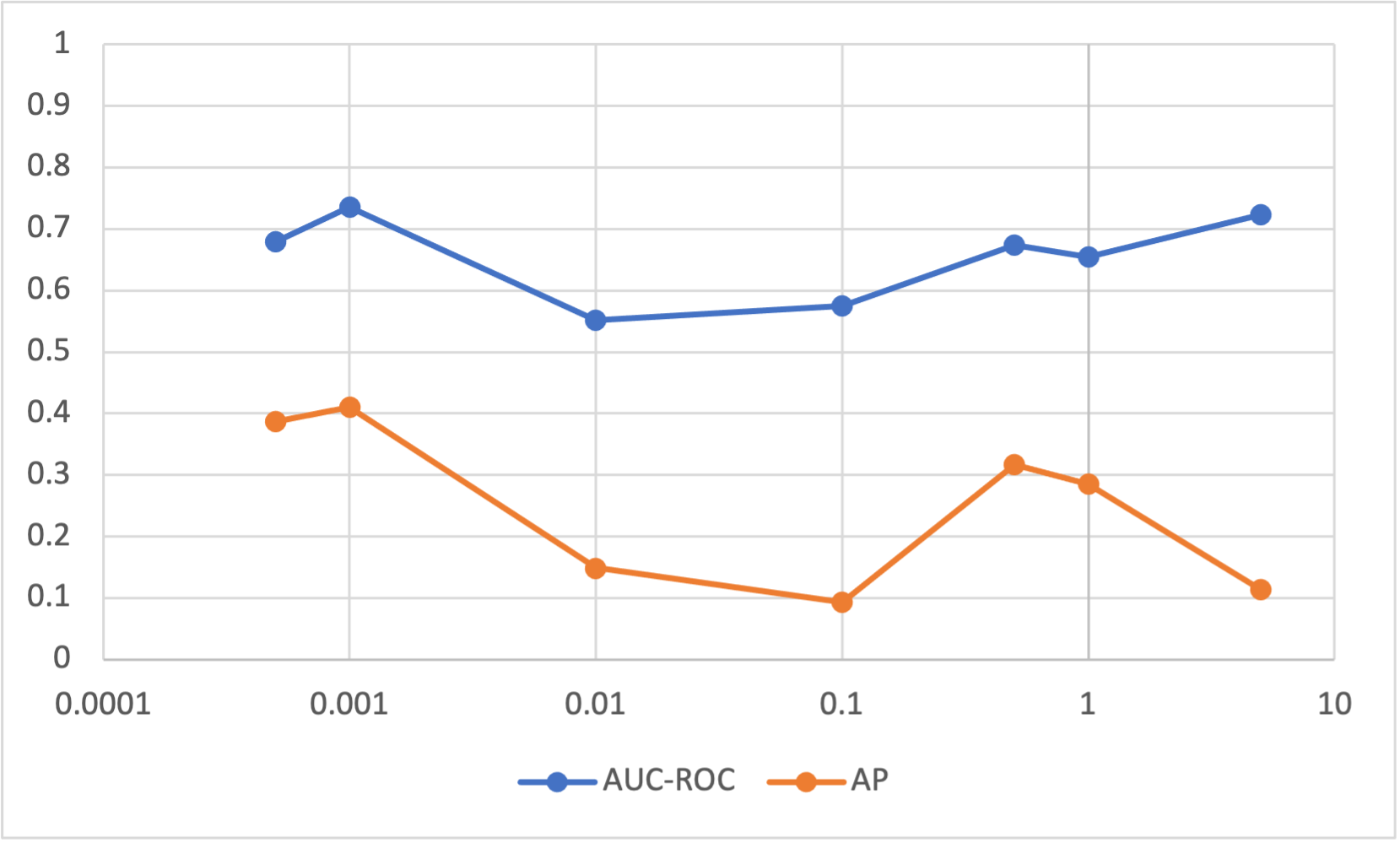}
		\caption{DeepSAD}
		\label{fig:hai_deepsad}
	\end{subfigure}
	\caption{Sensitivity of additional loss term weight  $\alpha$ on model performance with HAI data}
	\label{fig:sensitivity_hai}
\end{figure*}

\subsubsection{Summary}
The main difference between the proposed approach and DeepSAD is that the training contains an AR loss component. This 
loss 
component encourages the network to learn the normal dynamic relationship between future time steps and the past. In 
contrast, DeepSAD tries to train the network to map a window $T$ of  normal operation data to a hyper sphere. 
Although a network has 
the capacity to project a multivariate time series data into a hyper spherical space, it ignores the time series property and 
hence lacks the inherent induction bias that the AR learning formulation emphasizes, i.e., the  future can  be 
predicted 
from the past. Such induction bias aligns with the dynamic nature of the industrial time series data.  
Therefore, the direct projection representation learning from DeepSAD can be largely influenced by the loss term 
from labeled fault 
data, leading to large variation across random trials. Our proposed approach incorporates normality learning 
approach that takes into consideration of the time series nature, represented by the AR formulation, provides a better 
solution. 

On the other hand, we have noticed that the proposed approach did not improve upon normal data only model on 
HAI 
dataset. In HAI dataset, all the attack scenarios are different from each other. Therefore, the inherent characteristics 
of those 
anomalies are different. This is in contrary to the TEP dataset, where the training data contains all the fault types. In 
the TEP setup,  
a small number of simulation runs from each fault type are included  in the training data. However, the domain that 
represents 
anomalous behavior has shifted from training to testing in the HAI setup. This could explain the lackluster 
performance from HAI dataset in our experimental results. This also 
implies that the proposed approach can improve the anomaly detection performance on fault types for which we have 
labeled data for training, but it might have little impact on unseen fault types. The experimental results in 
our work motivates continued research in this area. We encourage future research work to address problems in such a 
setting. 

\subsection{Sensitivity Analysis}

Put the base model architecture aside, the only important hyper parameter of the proposed approach is the weight 
$\alpha$ on the additional loss 
term as in Equation \ref{eq:loss_c_all} and \ref{eq:loss_m_all} . We conducted a simple search on $\alpha$ by 
setting it to a set values $\{5.0, 1.0, 0.5, 0.1, 1\mathrm{e-}2, 1\mathrm{e-}3, 5\mathrm{e-}4\}$. We performed 
the same weight parameter search 
on DeepSAD. This weight sensitivity is graphed in Figure \ref{fig:sensitivity_tep} and \ref{fig:sensitivity_hai} for 
TEP and HAI dataset respectively. Both Auxiliary-LSTM and Margin-LSTM showed stable performance between 
$\alpha \in [0.5, 1]$ on both datasets.  On the other hand, we observed a large variation along the search range for 
DeepSAD. As mentioned in the previous section, we think the induction bias from an AR formulation could alleviate 
the variance impact caused by a small number of faulty samples during the training process. 

\section{Conclusion}
In this paper, we proposed a new approach to leverage the presence of a few anomalous cases on top of a large amount 
of normal operation data, an often encountered situation in industrial multivariate time series anomaly detection. We 
compared our approach with state of the art DeepSAD method, and demonstrated advantages over DeepSAD on two 
benchmark datasets. We demonstrated the importance of incorporating the dynamic nature of time series data to 
make the normal representation learning more stable and effective.  Our experimental results also revealed the 
limitation of the current approaches in this research area, i.e. the 
limited ability to improve detection sensitivity on anomaly types beyond those presented in the training data. We 
hope the reported results will motivate future research work to address this  problem setting, which is very relevant 
in real-world applications. 


\section*{Acknowledgment}
This material is based upon work supported by the Department of Energy, National Energy Technology Laboratory 
under Award Number DE-FE0031763.\footnote{Disclaimer:  This report was prepared as an account of work 
sponsored by an agency of the United States Government.  Neither the United States Government nor any agency 
thereof, nor any of their employees, makes any warranty, express or implied, or assumes any legal liability or 
responsibility for the accuracy, completeness, or usefulness of any information, apparatus, product, or process 
disclosed, or represents that its use would not infringe privately owned rights.  Reference herein to any specific 
commercial product, process, or service by trade name, trademark, manufacturer, or otherwise does not necessarily 
constitute or imply its endorsement, recommendation, or favoring by the United States Government or any agency 
thereof.  The views and opinions of authors expressed herein do not necessarily state or reflect those of the United 
States Government or any agency thereof.}
\bibliographystyle{IEEEtran}
\bibliography{deep_learning, SSAD}

\end{document}